# Probabilistic Surfel Fusion for Dense LiDAR Mapping

Chanoh Park[1,2], Soohwan Kim[1], Peyman Moghadam[1,2], Clinton Fookes[2] and Sridha Sridharan[2]

[1]Autonomous Systems Laboratory, CSIRO Data61, Brisbane, Australia
[2]Queensland University of Technology, Brisbane, Australia,
{Chanoh.Park, Soohwan.Kim, Peyman.Moghadam}@data61.csiro.au
{c.fookes, s.sridharan}@qut.edu.au

## Abstract

*With the recent development of high-end LiDARs, more and more systems are able to continuously map the environment while moving and producing spatially redundant information. However, none of the previous approaches were able to effectively exploit this redundancy in a dense LiDAR mapping problem. In this paper, we present a new approach for dense LiDAR mapping using probabilistic surfel fusion. The proposed system is capable of reconstructing a high-quality dense surface element (surfel) map from spatially redundant multiple views. This is achieved by a proposed probabilistic surfel fusion along with a geometry considered data association. The proposed surfel data association method considers surface resolution as well as high measurement uncertainty along its beam direction which enables the mapping system to be able to control surface resolution without introducing spatial digitization. The proposed fusion method successfully suppresses the map noise level by considering measurement noise caused by laser beam incident angle and depth distance in a Bayesian filtering framework. Experimental results with simulated and real data for the dense surfel mapping prove the ability of the proposed method to accurately find the canonical form of the environment without further post-processing.*

## 1. Introduction

In recent years, LiDAR-based Simultaneous Localization and Mapping (SLAM) has reached a significant level of maturity in many applications such as autonomous vehicles [6], UAVs (Unmanned Aerial Vehicles) [17] and 3D mobile mapping devices [24]. However, most of the LiDAR-based SLAM methods focus on trajectory estimation, and thus produce point clouds by aggregating LiDAR points. Due to the noise in measurements and errors in trajectory estimation, those point clouds suffer from the blurring effects and require batch post-processing to obtain consistent point clouds.

In this paper, we propose a new on-the-fly approach to align LiDAR scans from multiple views and merge point clouds using probabilistic surfel fusion. We model uncertainties of surfels based on incident angles, ranges and neighboring points and use them to find and merge correspondences. Moreover, the proposed data association that considers geometrical relationship between surfels offers a way to increase map surficial resolution without introducing further noise to the map. For the map presentation, we propose dual surfel maps, a sparse ESM (ellipsoid surfel map) and a dense DSM (disk surfel map) to take advantage of both surfel representations. ESM is a sparse 3D surfel map for localization, while DSM is a dense 3D surfel map to build accurate, much higher quality point clouds. We evaluate the performance of our method on both simulated and real data, and show that our method produces more accurate point clouds compared with the current state-of-the-art work.

The rest of the paper is organized as follows. In Section 2, we review related work on map generation and fusion methods. In Section 3 and 4, we describe the overview of our approach and details of our surfel extraction, matching and fusion method. We demonstrate our proposed method on simulated and real datasets in Section 5, and conclude the paper with future work in Section 6.

## 2. Related Work

Since the very beginning of SLAM [20], feature-based SLAM has been the dominant player in the community for the sake of simplicity and relatively lower computational complexity. Initial SLAM approaches in early 2000s utilized geometrical features of 2D LiDAR scanning such as corners and edges [13] which are tracked and updated as map elements. This concept was soon extended to 3D features [21]. However, despite the advantages of distinctive features, the feature-based SLAM approaches were limited to trajectory estimation as the features are too sparse to represent the 3D environment.

On the other hand, dense point clouds have also been popular with the advent of affordable 3D range sensors. They are useful not only for 3D reconstruction of the environment, but also to be used for dynamic object handling [10] and obstacle avoidance [8]. For building dense maps, batch optimization [3, 15] has been the most common approach for LiDAR-based mapping systems. However, those approaches just focused on the registration quality between local maps, ignoring the advantages of multiple observations in different view points, which is the key to reduce the noise in the maps and local deformations. Moreover, batch optimization has a limitation on life-long mapping. Thus, the introduction of on-the-fly dense map fusion is necessary for the next generation of LiDAR mapping systems.

Data association is one of the most important components in map fusion. In conventional 2D or 3D Extended Kalman Filtering (EKF) SLAM, finding the closest points with the Mahalanobis distances based on the sensor uncertainty is the most common approach for the data association [4, 13]. However, when it comes to dense maps, determining the object surface resolution is not obvious. Thus, most of the previous approaches opted to discretize the space into voxels [7].

Once matching is established, the next is to fuse the measurements for updated estimation. The conventional EKF-SLAM [13, 21] augments landmark positions into the state vector and updates the mean and covariance for every iteration. However, its cost increases quadratically whenever new features are added to the map. Thus, this approach could not be extended to the dense SLAM problem. Keller et al. [10] proposed a dense fusion method by simplifying the Bayesian estimation from 3D to 1D. In their approach, each map elements are independently updated, making its computation much simpler than the EKF case. They also utilized radial distortion as an initial uncertainty parameter and reduced the uncertainty whenever the surfel is observed again. ElasticFusion [22] further extended the uncertainty as a function of sensor motion to consider the uncertainty caused by motion blur. However, those simplified Bayesian models are not appropriate for dense LiDAR mapping where the existence of degeneracy in map elements often causes slower convergence.

## 3. Map Representation and Alignment

In this section, we first explain our dual surfel map representation, and utilizing sparse ellipsoid surfel map for localization. The proposed dual map representation not only let the system localize the current pose robust and fast but also construct a dense surfel map. The pose estimation result from this stage is utilized to align and merge local disk surfels using Bayesian filtering in the following section.

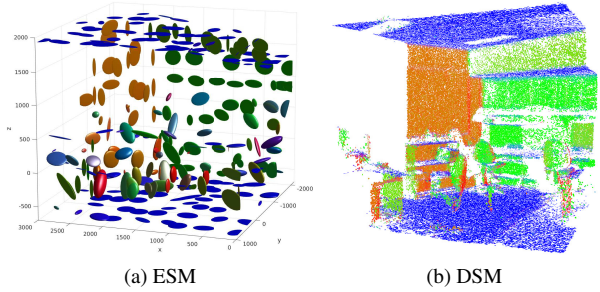

(a) ESM  (b) DSM

Figure 1. (a) Example of a 3D ellipsoid surfel map (ESM) with a 60 cm resolution and (b) a 2D disk surfel map (DSM) with a 1 cm resolution. Both are color-coded by normal directions. Recognize the ceiling and the floor in blue, and objects and walls in orange and green.

### 3.1. Dual Surfel Maps

We build two types of global surfel maps, ellipsoid surfel map (ESM) $\mathbb{S}_g$ and disk surfel map (DSM) $\mathbb{M}_g$. Each surfel map is individually updated with their local maps $\mathbb{S}_l$ and $\mathbb{M}_l$ which are extracted from the current laser scan. We assume that 3D point clouds without motion distortion are given by a 2D spinning laser [2] or multi-beam LiDAR such as Velodyne [14].

ESM consists of 3D ellipsoids extracted from laser points using multi-resolution voxel hashing [2]. Each ellipsoid is defined with a centroid $\mathbf{c} \in \mathbb{R}^3$ and a covariance matrix $\Sigma_\mathbf{c} \in \mathbb{R}^{3\times 3}$ which represent the distribution of points within the voxel. On the other hand, DSM is composed of 2D disk surfels $\varphi \in \mathbb{M}$ of which positions $\mathbf{p} \in \mathbb{R}^3$ are uniformly sampled from the laser points, and normal vectors $\hat{\mathbf{n}} \in \mathbb{R}^3$ are extracted from their neighboring points. In contrast to conventional surfels [10, 22], we associate uncertainties $\Sigma_\mathbf{p}, \Sigma_{\hat{\mathbf{n}}} \in \mathbb{R}^{3\times 3}$ with the position and normal vector of each disk surfel, which are later used to merge surfels based on Bayesian filtering. Figure 1 depicts an example of ESM and DSM. Note that 3D ellipsoid surfels are expressed with ellipsoids of their covariance matrices, while 2D disk surfels are expressed with disks with normal direction.

First, we utilize the multi-resolution ellipsoid surfel features [2] for localization in ESM. Note that in general ESM is too sparse to represent the environment [5, 19]. Thus, one might generate a point cloud by transforming all the laser points with estimated laser trajectories [2]. However, this method often produces blurred point clouds due to the noise in observations and errors in trajectory estimation. That is the reason we employ another surfel map, DSM [10, 22], which is much denser than ESM. While we localize the input point clouds with ESM, we sequentially update DSM to eliminate the noise and produce more accurate point clouds.

In summary, ESM is faster and more robust to run localization, and DSM is denser and more accurate to 3D reconstruct the environment.

### 3.2. Localization

We apply the point-to-plane ICP (Iterative Closest Point) to align local maps $\mathbb{S}_l$, $\mathbb{M}_l$ with respect to the global maps $\mathbb{S}_g$, $\mathbb{M}_g$. After finding the correspondences between local and global ESMs $\mathbb{S}_l$, $\mathbb{S}_g$ as in [2], we register the current local map $\mathbb{S}_l$ by minimizing the point-to-plane distances,

$$e = \sum_{i=1}^{n} e_i^2, \quad e_i = \hat{\mathbf{n}}_i^\top (\mathbf{p}_i^g - (\mathbf{R}\mathbf{p}_i^l + \mathbf{t})) \quad (1)$$

where $\mathbf{R} \in SO(3)$ and $\mathbf{t} \in \mathbb{R}^3$ are the rotation matrix and translation vector, $(\mathbf{p}_i^g, \hat{\mathbf{n}}_i^g) \in \mathbb{S}_g$ and $(\mathbf{p}_i^l, \hat{\mathbf{n}}_i^l) \in \mathbb{S}_l$ denote positions and normal vectors of the $i$-th correspondences among $n$ surfel matches, and $\hat{\mathbf{n}}_i = (\hat{\mathbf{n}}_i^g + \hat{\mathbf{n}}_i^l)/|\hat{\mathbf{n}}_i^g + \hat{\mathbf{n}}_i^l|$. We apply the Gaussian-Newton method that iteratively updates the rotation matrix and translation vector as

$$\mathbf{R}' = e^{[\delta \mathbf{r}]_\times} \mathbf{R}, \quad \mathbf{t}' = \mathbf{t} + \delta \mathbf{t} \quad (2)$$

where $\delta \mathbf{r}, \delta \mathbf{t} \in \mathbb{R}^3$, and $[\cdot]_\times$ denotes a skew-symmetric matrix. We also utilize a pose prior from the local trajectory estimation in a canonical form, $\xi \sim \mathcal{N}(\mu_\xi, \Sigma_\xi)$, where $\xi, \mu_\xi \in \mathbb{R}^6$, $\Sigma_\xi \in \mathbb{R}^{6 \times 6}$. Finally, we apply iteratively reweighted least squares with a t-distribution weight on residuals as in [1, 11]. Then, the normal equations can be written as

$$\left(\sum_{i=1}^{n} \frac{w_i}{\lambda_i} \mathbf{H}_i^\top \mathbf{H}_i + \Sigma_\xi^{-1}\right) \delta \xi = -\sum_{i=1}^{n} \frac{w_i}{\lambda_i} \mathbf{H}_i^\top e_i - \Sigma_\xi^{-1}(\bar{\xi} - \mu_\xi) \quad (3)$$

where $\delta \xi = (\delta \mathbf{r}^\top, \delta \mathbf{t}^\top)^\top$, $\bar{\xi}$ denotes the pose in the previous iteration, $\mathbf{H}_i = \partial e_i / \partial \delta \xi \in \mathbb{R}^{1 \times 6}$ is the Jacobian matrix, $w_i = (\nu + 1)/(\nu + (e_i/\sigma)^2)$ denotes the M-estimator weight, and $\nu$ and $\sigma^2$ are the number of degrees of freedom and variance of the t-distribution, respectively. Note that $\lambda_i^{-1}$ is introduced to penalize non-planar surfels, where $\lambda_i$ represents the smallest eigenvalue of $\Sigma_i^g + \Sigma_i^l + \Sigma_0$, and $\Sigma_0$ denotes the system noise. As the registration between local ESM $\mathbb{S}_l$ and global ESM $\mathbb{S}_g$ is based on point-to-plain ICP, we put more importance on the planer ellipsoid surfels. Thus, ellipsoid surfels integration between $\mathbb{S}_l$ and $\mathbb{S}_g$ after the optimization is simply defined by switching the global surfel with a new surfel when the new surfel has larger $\lambda_1$, $\lambda_2$ and smaller $\lambda_3$. The localization result is also applied to DSM $\mathbb{M}_l$ for a fusion in the next section.

## 4. Dense Surfel Matching and Fusion

When building the dense surfel map, there are two main issues with surfels extracted from LiDAR point clouds,

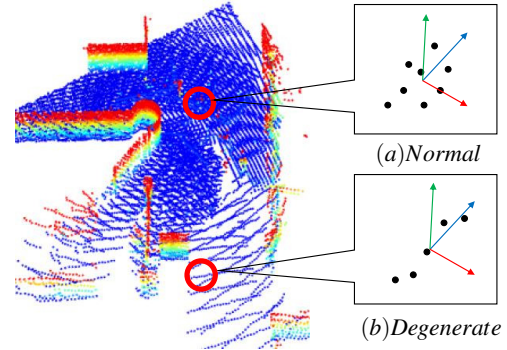

(a) Normal

(b) Degenerate

Figure 2. Example of LiDAR point clouds (right). Degenerate normal vector issue (left) when the normal (axis in green) is calculated from its neighboring points (black dots). (a) Desired shape of neighboring points. (b) Example of degenerate configuration which often occurs when scanned object is relatively far from the sensor. Estimated normal vector from this points set is not reliable.

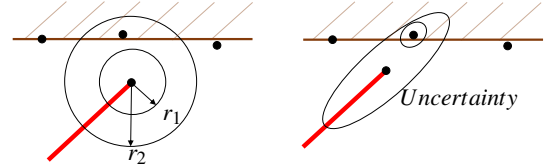

Figure 3. Conventional approaches to find point correspondences. Red lines denote rays and black dots depicts laser hit points. [left] Search by radius. [right] Search in the ray direction, considering the uncertainty in measurement.

compared with surfels generated from RGB-D point clouds. The first issue is surfel degeneracy. In the case of RGB-D data, it is obvious to find neighboring points in the image space. However, in the case of LiDAR data, neighboring points should be found in a 3D space based on distance (Figure2(a)), and thus the chance of getting degenerate surfels is relatively high as shown in Figure 2(b). This occurs quite often when objects are far from the sensor, and the scanning line pattern appears on surface. To address this issue, we model uncertainties in positions and normal vectors for each surfel in Section 4.1.

The other issue is that surfel matching is not straightforward. In the case of RGB-D surfels, projective data association [22] is readily applicable. However, it is not the case with LiDAR surfels because there is no projection plane. The simplest way is to find the one with the closest distance [15] as shown in the left of Figure 3. However, if the search radius is smaller than the sensor noise (e.g. $r_1$ in the left of Figure 3), the matching accuracy drops, whereas in the opposites case (e.g. $r_2$ in the left of Figure 3), we get a lower map resolution. Considering that the depth uncertainty of a moderate LiDAR is high along its beam direction, the radius search method severely reduces the map resolution. One might consider uncertainty in measurement as shown in the right of Figure 3 which is common in the filtering-based SLAM [4]. However, in this case it is tricky to control the map resolution without discretizing the environment [7]. To address this issue, we propose a new algorithm for surfel matching that preserves the map resolution in Section 4.2.

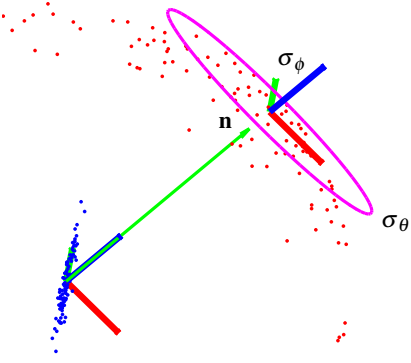

Figure 4. An example of a degenerate surfel normal (green line) which is estimated from the neighbouring points along an edge (blue dots). The uncertainty in the normal vector is described with end tip points of random samples (red dots) and represented with an ellipse (magenta circle) on its tangential space.

### 4.1. Surfel Uncertainty Modeling

**Uncertainty in Position** For surfel centroid uncertainty, we utilize the sensor noise model proposed in [18]. As the LiDAR point clouds are directly used as surfel centroids, they have the same uncertainty characteristics as LiDAR measurement which are governed by incident angle, ambient temperature and humidity [12]. As depicted in the right of Figure 3, the positional uncertainty of a surfel is high along the beam direction. We model this uncertainty in the surfel position with an ellipsoid with three principal axes. The amount of uncertainty along the beam direction is defined by the sum of the distance uncertainty $\sigma_r^2$ and additional uncertainty $\sigma_i^2$ caused by the incident angle [12, 16]. Thus, the complete uncertainty of each surfel position in the world coordinate is given as

$$\Sigma_\mathbf{p} = {}^w\mathbf{R}_l {}^l\mathbf{R}_b \Sigma_b ({}^w\mathbf{R}_l {}^l\mathbf{R}_b)^\top \quad (4)$$

where ${}^w\mathbf{R}_l$ and ${}^l\mathbf{R}_b$ are rotation matrices from laser to world coordinates, and from beam to laser coordinates, respectively. The uncertainties in the beam coordinates $\Sigma_b = diag(\sigma_x^2, \sigma_y^2, \sigma_z^2)$ is modeled as in [18] with additional uncertainties $\sigma_i^2, \sigma_r^2$ added to $\sigma_z^2$. Note that each surfel centroid has independent uncertainty according to the beam source and different sensor locations.

**Uncertainty in Normal Vector** The uncertainty of the normal vector is directly related to the three eigenvalues ($\lambda_1 \geq \lambda_2 \geq \lambda_3$) of the covariance matrix calculated from its neighboring points. The following cases yield unstable or incorrect normal vectors; (1) $\lambda_1$ and $\lambda_2$ are too small (particles), (2) $\lambda_1 \gg \lambda_2$ (edges), (3) $\lambda_3$ is too large (blobs).

Note that the tip of a normal vector moves on a unit sphere. Thus, its uncertainty has only two degrees of freedom. As the uncertainty propagation on a manifold space is not easy to define, we propose an approximation model which defines two degrees of freedom uncertainty in a tangential space at the tip of the normal vector as shown in Figure 4.

To reflect the relationship between the shape and uncertainty, we define the uncertainty of the normal direction $diag(\sigma_\theta, \sigma_\phi)$ on the tangential space in $\mathbb{R}^2$ as a function of eigenvalues. We start with defining uncertainty attributes: $\alpha_\theta = a\lambda_1^{-1} - 0.5$, $\alpha_\phi = b\lambda_2^{-1} - 0.5$, $\alpha_{z1} = log(\lambda_3/\lambda_1)c + 0.5$, $\alpha_{z2} = d\lambda_3 - 0.5$. Note that $\alpha_\theta$ and $\alpha_\phi$ penalize too small variances along the first two principal axes, while $\alpha_{z1}$ and $\alpha_{z2}$ penalize too large relative and absolute variances along the third axis, respectively. Here, $a, b, c, d$ are scaling coefficients and are determined statistically. As a result, we model the uncertainty in a normal vector on a tangential space by integrating the attributes in a sigmoid function as

$$\sigma_\theta = (1 + e^{-w(\alpha_\theta + \alpha_{z1} + \alpha_{z2})})^{-1} \quad (5)$$
$$\sigma_\phi = (1 + e^{-w(\alpha_\phi + \alpha_{z1} + \alpha_{z2})})^{-1} \quad (6)$$

where $w$ is a scaling factor for the sigmoid function. Finally, the uncertainty in a normal vector in the world coordinates is defined as

$$\Sigma_\mathbf{n} = \mathbf{v} diag(\sigma_\theta, \sigma_\phi, \varepsilon) \mathbf{v}^\top \quad (7)$$

where $\varepsilon$ is added to prevent a singularity problem in matrix inversion. The Eigen vector matrix $\mathbf{v}$ is utilized to align the normal uncertainty direction with the underlying neighboring points shape in the world coordinate frame.

### 4.2. Surfel Matching

This section describes our method for finding matched surfels between the global DSM $\mathbb{M}_g$ and local DSM $\mathbb{M}_l$ for surfel fusion in the following section. Once the transformation is decided by ICP, the local surfels $\mathbb{M}_l$ are transformed into the world coordinates and ready to find the matched surfels in the global dense surfel map $\mathbb{M}_g$. The matching process begins with finding a set of candidate surfels $\mathbb{A}_g$ for each surfel $\varphi_l \in \mathbb{M}_l$. For efficient matching, initial matching candidates are selected by an octree-based nearest neighbor search algorithm. Then, the resolutional distance $r$ between each source and destination pair in Figure 5 is compared with a resolution threshold $\theta_r$ to decide if their projections are close enough on the surface. If so, we check the depth

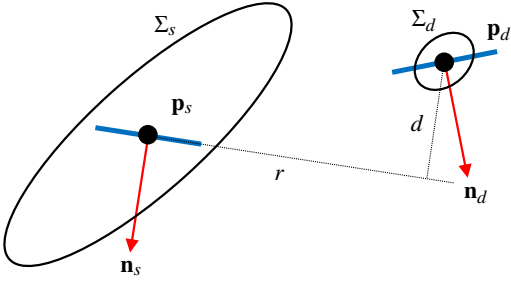

Figure 5. Illustration of the resolutional distance $r$ and the depth distance $d$ in surfel matching. Blue bars represent the side view of surfels, while red arrows denote normal vectors. The resolution threshold $\theta_r$ to which $r$ is compared decides the map surface resolution whereas the projective matching distance threshold $\theta_d$ to which $d$ is compared is related to how deep it will search for the matching.

---

**Algorithm 1** Finding Surfel Matches

**Input:** Global DSM $\mathbb{M}_g$ and a surfel $\varphi_l \in \mathbb{M}_l$
**Output:** A set of matched surfels $\mathbb{B}_g \subseteq \mathbb{M}_g$
$\mathbb{A}_g \leftarrow OctreeSearch(\varphi_l, \mathbb{M}_g)$
**foreach** $\varphi_g \in \mathbb{A}_g$ **do**
    $[r, d] \leftarrow Point2PlaneDist(\varphi_g, \varphi_l)$
    **if** $r < \theta_r$ **then**
        $\sigma^2 = \hat{\mathbf{n}}_s^\top \Sigma_s \hat{\mathbf{n}}_s + \hat{\mathbf{n}}_d^\top \Sigma_d \hat{\mathbf{n}}_d$
        **if** $d/\sigma < \theta_d$ **then**
            $\mathbb{B}_g \leftarrow \mathbb{B}_g \cup \varphi_g$
        **end**
    **end**
**end**

---

distance $d$ in the Mahalanobis distance. To consider the uncertainty only along the normal direction, we propagate the positional 3D uncertainties of source and destination surfel $\Sigma_d, \Sigma_s$ into 1D along each normal direction by $\sigma^2 = \hat{\mathbf{n}}^\top \Sigma \hat{\mathbf{n}}$. Finally, if the 1D Mahalanobis distance along the surface normal direction is less than a threshold $\theta_d$, we assume that they are correspondences, and put the matched surfel into $\mathbb{B}_g$. Note that the resolutional distance $d$ is compared in Euclidean space to preserve the desired surface resolution in Euclidean space. Algorithm 1 summarizes this surfel matching process. Note that our matching method enables the matching process to search more along the beam direction while effectively maintaining the desired surface resolution without a voxel grid.

### 4.3. Surfel Fusion

In this section, we describe our method of fusing surfel matches. We start by defining the position and normal vector of a surfel $\varphi$ as random variables $\mathbf{p} \sim \mathcal{N}(\mu_{\mathbf{p}}, \Sigma_{\mathbf{p}})$, $\hat{\mathbf{n}} \sim \mathcal{N}(\mu_{\hat{\mathbf{n}}}, \Sigma_{\hat{\mathbf{n}}})$. Given a surfel observation $\varphi_l \in \mathbb{M}_l$ and the matched surfel $\varphi_g \in \mathbb{M}_g$, our objective is to find the better estimation. Assuming two observations are independent, the Kalman filter gives a Bayesian update formula as

$$\mu'_g = \Sigma'_g (\Sigma_g^{-1} \mu_g + \Sigma_l^{-1} \mu_l) \qquad (8)$$
$$\Sigma'_g = (\Sigma_g^{-1} + \Sigma_l^{-1} + \Sigma_s^{-1})^{-1} \qquad (9)$$

which can immediately be applied for the fusion of centroids as listed in the second and third line of Algorithm 2.

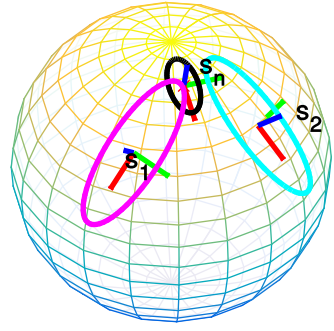

Figure 6. Normal direction uncertainties of two surfel normals (magenta, crayon) on a unit sphere and its fusion (black).

However, a different approach should be taken for normal vectors as it is in a manifold. As we described in Section 4.1 the canonical form of normal uncertainty lies on the tangential space of the unit sphere. Thus, to handle the uncertainty propagation, we lift the 2D normal uncertainty to a 3D space by Equation (7) and fuse them in a 3D space as Line 13, 14 in Algorithm 2. An example of two surfel normal fusion on the unit sphere is depicted in Figure 6. Generally, the propagated uncertainty $\Sigma'_{\mathbf{n}_d}$ is not tangential to the sphere surface. Thus, tangentiality should be reinforced after the propagation by decomposing $\Sigma'_{\mathbf{n}_d}$ and forcing its z axis to be aligned with the fused normal direction $\mathbf{n}'_d$ (Line 15 to 18 in Algorithm 2). As this is a linearized method, we limit its application to the situations where the distance of two vectors on the surface is small enough.

There are some cases where the underlying original point geometry of a surfel is degenerate as depicted in Figure 2 where the approximated normal fusion model cannot handle this. A surfel degeneracy can be easily found by looking at uncertainty ratio of $\sigma_\theta, \sigma_\phi$. When it is degenerate, the uncertainty of the first principal axis $\sigma_\theta$ is far higher than $\sigma_\phi$. Instead of just throwing those surfels away, we keep them in the surfel pool as a degenerate surfel and wait until it is properly observed. When one of the target or source surfels is degenerate, the new normal direction and uncertainty follows the ordinary one. In case both of the source and destination surfels are degenerate, the normal is acquired by the cross product of the first principal directions.

Note that the added system uncertainties $\sigma^s_\theta, \sigma^s_\phi, \sigma^s_{ICP}$

**Algorithm 2** Surfel Fusion

**Input:** Map surfel $\varphi_i^{tar}$, Input surfel $\varphi_i^{src}$
**Output:** Updated map surfel $\varphi_i$
**foreach** *pair of matched surfels $\varphi_i^{tar}, \varphi_i^{src}$* **do**
    Centroid Fusion:
    $\Sigma_d \leftarrow (\Sigma_s^{-1} + \Sigma_d^{-1} + \Sigma_s^{-1})^{-1}$
    $\mathbf{p}_d \leftarrow (\Sigma_s^{-1} + \Sigma_d^{-1})^{-1}(\Sigma_d^{-1}\mathbf{p}_d + \Sigma_s^{-1}\mathbf{p}_s)$
    **if** $\sigma_\theta \gg \sigma_\phi$ **then**
        surfel is degenerate
    **end**
    **if** $\varphi_i^{tar} = degenerate, \varphi_i^{src} = degenerate$ **then**
        $\mathbf{n}'_d \leftarrow \mathbf{v}_s \times \mathbf{v}_d$
    **else if** $\varphi_i^{tar} = degenerate$ **then**
        $\varphi_i \leftarrow \varphi_i^{src}$
    **else if** $\varphi_i^{src} = degenerate$ **then**
        $\varphi_i \leftarrow \varphi_i^{tar}$
    **else**
        Normal Direction Fusion:
        $\Sigma'_{\mathbf{n}_d} \leftarrow (\Sigma_{\mathbf{n}_s}^{-1} + \Sigma_{\mathbf{n}_d}^{-1})^{-1}$
        $\mathbf{n}'_d \leftarrow \Sigma'_{\mathbf{n}_d}(\Sigma_{\mathbf{n}_s}^{-1}\mathbf{n}_s + \Sigma_{\mathbf{n}_d}^{-1}\mathbf{n}_d)$
        $[\lambda \quad \mathbf{v}] \leftarrow SVD(\Sigma'_{\mathbf{n}_d})$
        $\Sigma_{new} \leftarrow \lambda + diag(\sigma_\theta^s, \sigma_\phi^s, -\lambda_3)$
        $\mathbf{R} \leftarrow [u_1 \times \mathbf{n}'_d \quad (u_1 \times \mathbf{n}'_d) \times \mathbf{n}'_d \quad \mathbf{n}'_d]$
        $\Sigma'_{\mathbf{n}_d} \leftarrow \mathbf{R}\Sigma_{new}\mathbf{R}^T$
    **end**
**end**

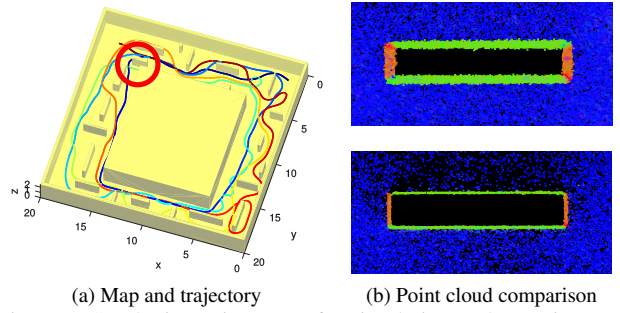

(a) Map and trajectory     (b) Point cloud comparison

Figure 7. Synthetic environment for simulation and experimental results with simulated data. (a) A spinning LiDAR scans an office-like environment of $20 \times 20$ m several times, (b) Top views of the red circle area in (a) to compare point clouds generated by CT-SLAM [2] (top) and our method (bottom). Points are color-coded by normal directions.

prevent the surfels from being over-fitted to repeated systematic errors such as noise points caused by a mixed pixel problem [23].

The surfels that are not matched to the global map will be added to the global map as a new unstable surfel. To effectively remove the surfels generated from LiDAR non-Gaussian noise, any unstable surfels that are not observed for a certain period of time (*e.g.* 5 min) when the sensor revisited the surfel within a certain radius will be deleted from the global map.

## 5. Experimental Results

### 5.1. Simulated Data

In order to evaluate the proposed method, we compare the fused map from a simulation data with its ground truth. We first generated synthetic data with a LiDAR simulator. As shown in Figure 7(a), a spinning LiDAR simulates hit points using ray casting with measurement noise of $\sigma = 15$ mm along the beam direction in an office-like environment of $20 \times 20$ m. Then, the simulated data is used to build a fused map by the proposed method. Finally, the position and normal errors of each surfels can be calculated as its ground truth is known from the simulation process.

Table 1 describes the simulated result in more detail by comparing the errors in positions and normal vectors between CT-SLAM [2] and our proposed method. The positional errors are calculated by point-to-plane distances, while the normal vector errors are calculated as an angle difference between mesh normals and map normals. Our proposed probabilistic surfel fusion produced an accurate point cloud with about 3 times less errors in both positions and normal vectors. Also, as shown in Figure 7(b), our method produced sharper and cleaner point clouds on the wall than CT-SLAM [2].

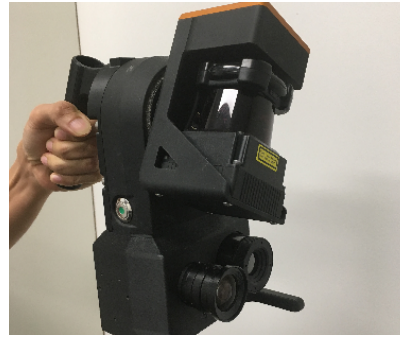

Figure 8. The experimental handheld 3D spinning LiDAR for mobile mapping. It contains a 2D laser, an IMU, an encoder, a color camera and a thermal camera.

|  | Size | Position Err. | | Normal Err. | |
| --- | --- | --- | --- | --- | --- |
|  |  | mean | std. | mean | std. |
| CT-SLAM | $4.9 \times 10^6$ | 10.6 | 78.7 | 9.5 | 10.6 |
| Our method | $2.6 \times 10^6$ | **3.7** | **7.7** | **3.2** | **7.3** |

Table 1. Comparison of point cloud accuracy between CT-SLAM [2] and our method given the same simulation data. The unit for errors is mm. Size is the total number of surfels in the final map.

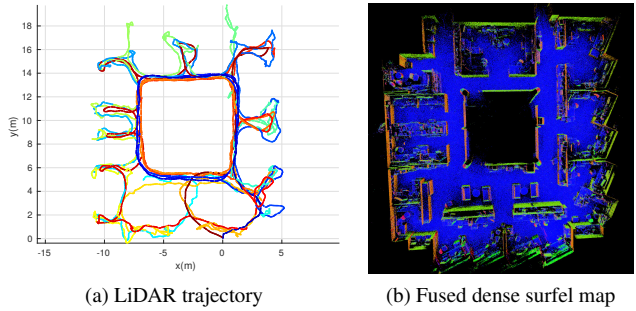

(a) LiDAR trajectory  (b) Fused dense surfel map

Figure 9. Real environment and experimental result with real data. (a) LiDAR trajectory in an office building of $20 \times 20$ m. The total trajectory length is 707 meters and recorded for 24 min. (b) Top view of the global dense surfel map generated by the proposed method. Note that the ceiling is removed to show the details inside. Color represents normal directions.

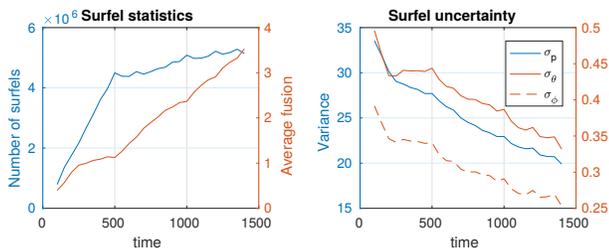

Figure 10. Surfel statistics and uncertainties. [left] The number of surfels and the average number of fusion per surfel, [right] Uncertainties of surfel positions and normal vectors.

## 5.2. Real Data

For the real data experiments, an experimental hand-held 3D spinning LiDAR system is used. The system consists of a spinning Hokuyo UTM-30LX laser, an encoder, a Microstrain 3DM-GX3 IMU, Grasshopper3 2.8 MP color camera and Optris PI 450 thermal-infrared 382 x 288 pixel camera (Figure 8). For the experimental results presented here, we only use the 3D spinning LiDAR, IMU and the encoder data.

To demonstrate the advantages of the proposed method, we compare our fused map from the proposed method with a map from the ordinary global batch optimization method (CT-SLAM [2]). Figure 9(a) shows the scanning trajectory in an office building of about $20 \times 20$ m. The data utilized in the comparison is collected by the sensor in Figure 8, while the operator is moving around the office at a speed of about 0.5 meters per second. We obtained local point clouds by 3D spinning LiDAR and local trajectory optimization using [2], and collected 1441 views in total for 24 minutes. Each of the views contains 194k 3D points. Figure 9(b) shows the top view of the sequentially fused dense surfel map.

Figure 10 shows surfel statistics and uncertainties changing over time. The number of surfels steeply increases until they fill the most of the space, and then slowly increases based on surfel fusion. The surfel count occasionally drops due to the removal of unstable surfels. Recognize that the average number of updates per surfel monotonously increases over time. On the other hand, the uncertainties of positions and normal vectors of surfels decrease as they are observed multiple times. Here, the positional uncertainties are calculated by the distance errors to the closest surfaces.

|  | Size | Noise | |
|---|---|---|---|
|  |  | mean | std. |
| CT-SLAM | $4.2\times10^6$ | 3.10 | 46.67 |
| CT-SLAM* | $2.1\times10^6$ | 0.41 | 10.21 |
| Our method | $2.1\times10^6$ | **0.22** | **7.86** |

Table 2. Comparison of noise in the point clouds generated by CT-SLAM [2] and our method given the same real data. CT-SLAM* denotes the CT-SLAM result with post-processing. Noise is the distances to the closest surfaces extracted from the point clouds. The unit for noise is mm. Size is the total number of surfels in the final map.

As the ground truth is not available for the real data, we extracted surface meshes from the point clouds by [9] and evaluated the distances to the closest surfaces as a metric for accuracy. Table 2 shows that our method produced much more accurate point cloud than CT-SLAM does. Our result is even better than the CT-SLAM point clouds after post-processing which performs outlier removal and k-nearest neighbor noise filtering. The reduced noise in our result is visualized in Figure 11.

Figure 12 compares dense surfel maps from CT-SLAM and the proposed method. The fused map from the proposed method shows noticeably a lesser noise level than the original point cloud. This is due to in the proposed method matched surfels are merged and unstable surfels are regularly removed during the fusion. The proposed method produced a sharp surfel map with less redundancy and noise.

## 6. Conclusion

In this paper, we proposed a new approach for dense LiDAR mapping by applying probabilistic surfel fusion. Particularly, we built dual surfel maps, 3D ellipsoid surfel map (ESM) and 2D disk surfel map (DSM). We aligned the point clouds based on sparse ESM, and updated dense DSM based on Bayesian filtering. In addition, we modeled uncertainties in positions and normal vectors for each surfel and considered degenerated surfels due to sparse laser hit points. Also, the proposed data association method increases surface resolution of the map while successfully suppressing noise level. Experimental results with both simulated and real data show that our method produces more accurate surfel maps with less noise and a minimum amount of map el-

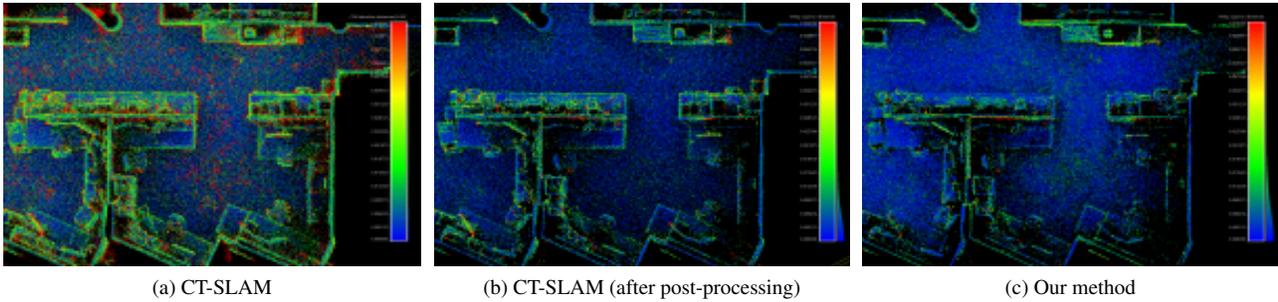

(a) CT-SLAM  (b) CT-SLAM (after post-processing)  (c) Our method

Figure 11. Comparison of noise in point clouds from the top view. Points are color-coded by noise, 0 (blue) to 5 mm (red). The points on the floor and tables contains more noise in CT-SLAM results than our result.

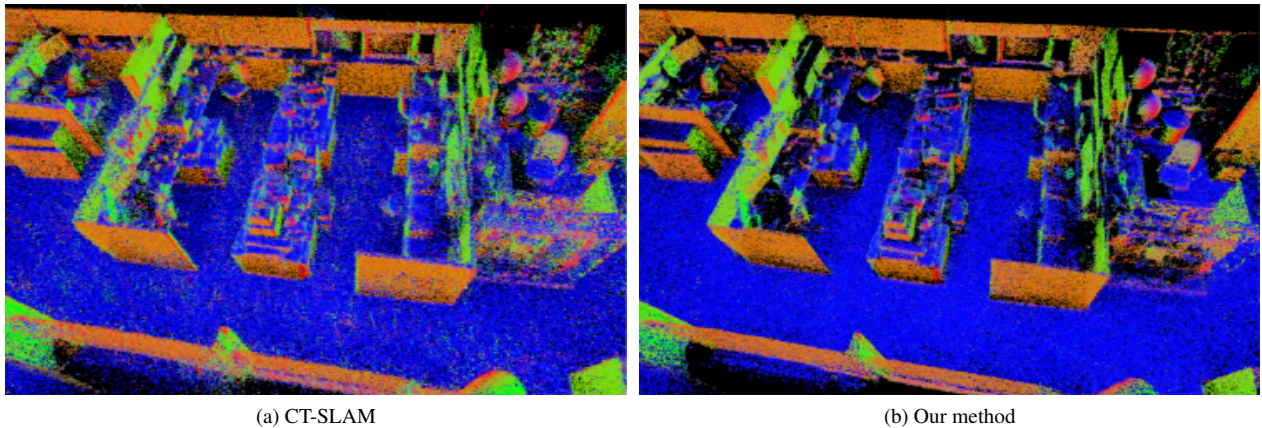

(a) CT-SLAM  (b) Our method

Figure 12. Comparison of dense surfel maps. The redundant surfels in CT-SLAM due to local deformation and the mixed pixel problem are clearly combined in our results, and the noise in normal vectors is dramatically reduced in our results.

ements, compared with the previous work. In future work, our method can be further extended to real-time LiDAR mapping with hardware and software optimization to produce an accurate dense LiDAR surfel map on the fly. This is applicable because our method sequentially updates surfel maps rather than globally optimizing the maps by batch processing.

**Acknowledgements**   The authors gratefully acknowledge funding of the project by the CSIRO and QUT. The institutional support of CSIRO and QUT, and the help of several individual members of staff in the Autonomous Systems Laboratory including Tom Lowe, Gavin Catt and Mark Cox are greatly appreciated.